\documentclass[conference]{IEEEtran}
\IEEEoverridecommandlockouts
\usepackage[T1]{fontenc}
\usepackage[utf8]{inputenc}
\usepackage{times}        % Times New Roman for body
\usepackage{courier}      % Courier for code
\usepackage{soul}
\usepackage{float}
\usepackage{url}
\usepackage{tabularx}
\usepackage{array} % required for text
\usepackage{makecell}
\usepackage{multirow}
\usepackage{booktabs}
\usepackage[colorlinks,urlcolor=blue,linkcolor=blue,citecolor=blue]{hyperref}
\usepackage[dvipsnames,table]{xcolor}
\usepackage{xspace}
\usepackage{rotating}
\usepackage{balance}
\usepackage{graphicx} % Required for \resizebox
\usepackage{adjustbox} % Required for adjustbox
\usepackage{comment}
\usepackage{cite}
\usepackage{amsmath,amssymb,amsfonts}
\usepackage{algorithmic}
\usepackage{cleveref}
\usepackage{listings}
\usepackage{enumitem}
\usepackage[most]{tcolorbox} % needed for the task example boxes

\definecolor{paretogreen}{RGB}{168,219,140}
\definecolor{paretoyellow}{RGB}{252,237,140}
\definecolor{LightCyan}{rgb}{0.88,1,1}

\def\BibTeX{{\rm B\kern-.05em{\sc i\kern-.025em b}\kern-.08em
    T\kern-.1667em\lower.7ex\hbox{E}\kern-.125emX}}

\newcommand{\prompt}{\mathcal{P}}
\newcommand{\glb}{\mathcal{G}}
\newcommand{\loc}{\mathcal{S}}
\newcommand{\completion}{\mathcal{C}}

\newcommand{\ner}{\mathcal{N}}
\newcommand{\knowledge}{\mathcal{K}}
\newcommand{\llm}{\mathcal{L}}
\newcommand{\fit}{FiT\xspace}

\begin{document}

\title{Find Before You Fine-Tune: A Diagnostic Study of Small LLMs for Cybersecurity QA
}

% ---- For double-blind review (current). Camera-ready author block is
%      commented out below. ----
% \author{
% \IEEEauthorblockN{Anonymous Author(s)}}

\author{
\IEEEauthorblockN{
Shaswata Mitra\IEEEauthorrefmark{1},
Subash Neupane\IEEEauthorrefmark{2},
Trisha Chakraborty\IEEEauthorrefmark{3},
Himanshu Tripathi\IEEEauthorrefmark{4},\\
Sudip Mittal\IEEEauthorrefmark{5},
Aritran Piplai\IEEEauthorrefmark{7},
Shahram Rahimi\IEEEauthorrefmark{6}
}
\IEEEauthorrefmark{1}\IEEEauthorrefmark{4}\IEEEauthorrefmark{5}\IEEEauthorrefmark{6}The University of Alabama -- \{\IEEEauthorrefmark{1}smitra3, \IEEEauthorrefmark{4}htripathi, \IEEEauthorrefmark{5}sudip.mittal, \IEEEauthorrefmark{6}shahram.rahimi\}@ua.edu\\
\IEEEauthorrefmark{2}Meharry Medical College -- \IEEEauthorrefmark{2}subash.neupane@mmc.edu\\
\IEEEauthorrefmark{3}Mississippi State University -- \IEEEauthorrefmark{3}tc2006@msstate.edu\\
\IEEEauthorrefmark{7}The University of Texas at El Paso -- \IEEEauthorrefmark{7}apiplai@utep.edu\\
}

\maketitle

\begin{abstract}
Large Language Models (LLMs) are increasingly fine-tuned for critical-domain Question-Answering (QA), yet choosing which small model to adapt, before paying the cost of adaptation, remains difficult. Fine-tuning can improve domain alignment, but it may also erode prior knowledge, weaken instruction-following, or increase hallucination, especially when labeled data are scarce or rapidly evolving as in cybersecurity. We present \fit (Find before Fine-Tune), a task-oriented diagnostic framework that characterizes small LLMs along three capabilities required for cybersecurity QA: vocabulary recognition, parametric knowledge, and contextualization of retrieved information. Using \fit, we conduct an empirical study of five open-weight 7-billion-parameter models under two fine-tuning regimes. We find that fine-tuning does not uniformly help: it consistently degrades vocabulary and parametric knowledge in small models, and the two regimes trade off differently. Knowledge-focused tuning causes moderate, \emph{rank-preserving} degradation, whereas instruction-focused tuning collapses measured knowledge through induced abstention, inverting the knowledge ranking while leaving retrieval-grounded contextualization essentially intact. We quantify these regime-specific patterns with rank-correlation analysis and show that pre-fine-tuning \fit scores anticipate the direction of post-tuning change. Our results suggest that task-oriented diagnosis can screen out unsuitable models, avoid unnecessary fine-tuning, and support safer deployment of small LLMs in cybersecurity QA pipelines.
\end{abstract}


% ############################################################################################################
\section{Introduction}
Large Language Models (LLMs) have produced remarkable advances in Natural
Language Processing (NLP), demonstrating unprecedented language understanding
and generation capabilities. Organizations across many industries now deploy
LLMs as domain-specific Question Answering (QA) systems. In cybersecurity, a field tied to critical infrastructure and national security, incorrect model responses can trigger attacks, facilitate breaches, or expose sensitive information, making reliable deployment especially consequential \cite{llm_mistakes}. Effective cybersecurity QA requires a model to possess domain vocabulary,
specialized knowledge, and the ability to synthesize heterogeneous retrieved
information. Consider the query: \emph{``Can the LangChain vulnerability affect my
system?''} Answering correctly demands both general knowledge of LangChain
vulnerability classes and context-specific awareness of the user's deployed
version. These two information types are qualitatively different, yet both are
necessary for a correct, non-leaking response.

Because cybersecurity knowledge evolves continuously and labeled data are scarce, practitioners frequently couple a small, deployable LLM with Retrieval-Augmented Generation (RAG) \cite{lewis2020retrieval} and, where data permit, fine-tune the model to the deployment scope. This raises a practical question that precedes deployment: \emph{which} small model should be selected, and \emph{whether} fine-tuning will help at all. Answering empirically by fine-tuning every candidate is expensive in both compute and time. We therefore ask whether a lightweight, pre-adaptation diagnosis can guide this choice.

\begin{figure*}[ht]
    \centering
    \includegraphics[width=0.9\textwidth]{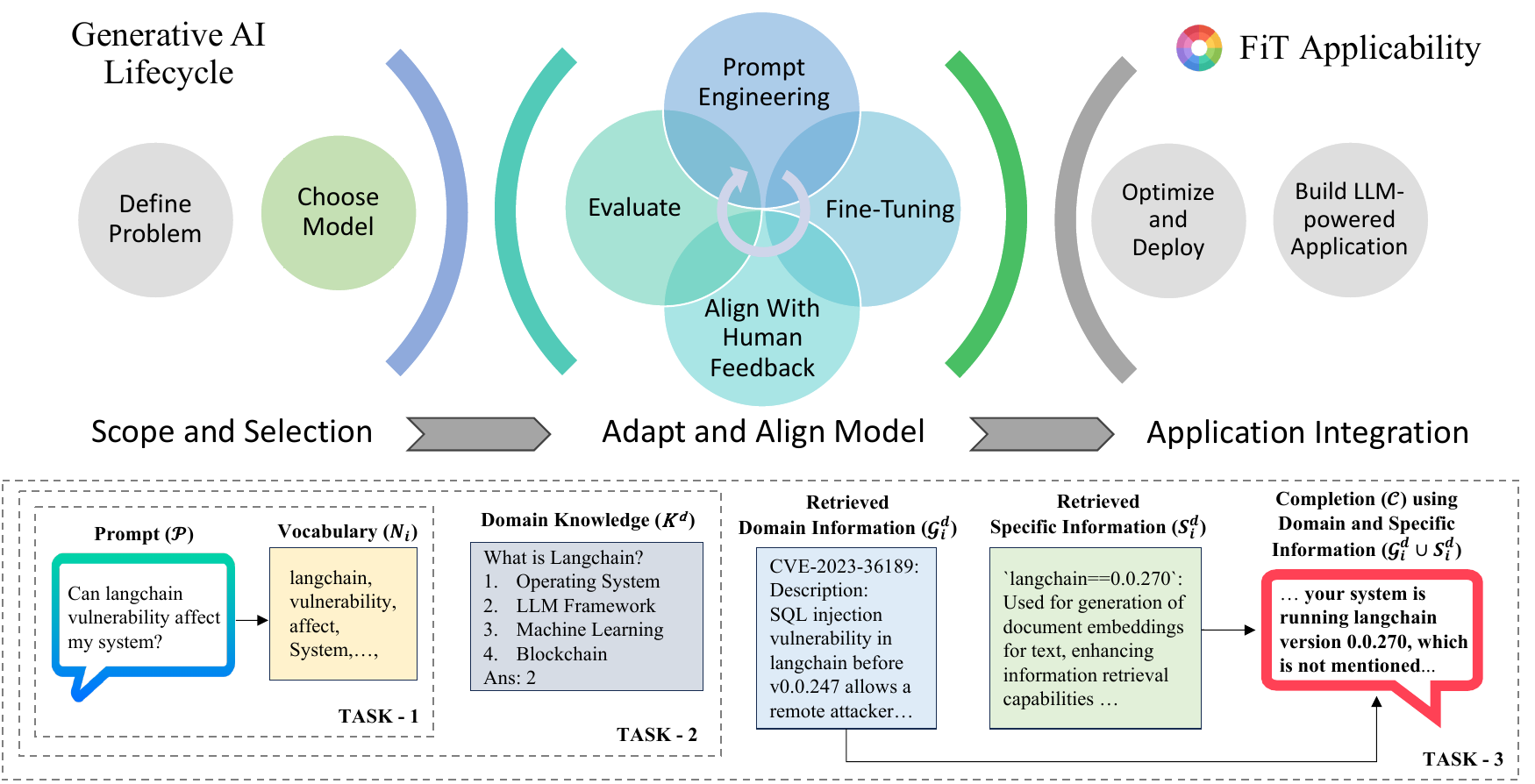}
    \caption{Implementation scope of \fit in the generative-AI life-cycle (colored area implies applicability) with an example of our three evaluation tasks. By aligning the tasks in a complete process, we visualize the propagation of the domain suitability required to generate Completion ($\completion$) for a given Prompt ($\prompt$).}
    \label{Fig:task_example}
\end{figure*}

We introduce \fit, a task-oriented diagnostic framework that characterizes a small LLM's domain understanding and contextualization ability for knowledge-intensive QA. Whereas existing cybersecurity benchmarks such as CYBERBENCH \cite{liucyberbench} and SecureBERT \cite{aghaei2022securebert} report static, pre-trained performance, to the best of our knowledge \fit is the first framework to diagnose a small LLM's cybersecurity QA suitability \emph{before} adaptation and to characterize how fine-tuning reshapes that suitability. Rather than positing \fit as a general predictive benchmark, we use it as a lens to study \emph{how} small models behave before and after two distinct fine-tuning regimes, so that organizations can screen candidates and anticipate the direction of post-tuning change. \fit decomposes suitability into three aspects (\textit{Vocabulary, Knowledge, Contextualization}), and we instantiate the study in cybersecurity\footnote{Code and data: github.com/shaswata09/FiT}. Concretely, we investigate three research questions:

\begin{itemize}[noitemsep,leftmargin=*]
    \item \textbf{RQ1.} Can a lightweight, pre-adaptation diagnosis characterize a small LLM's suitability for cybersecurity QA along vocabulary, knowledge, and contextualization, without the cost of fine-tuning each candidate?
    \item \textbf{RQ2.} How do knowledge-focused and instruction-focused fine-tuning regimes alter these capabilities in small LLMs, and do they help uniformly?
    \item \textbf{RQ3.} Do pre-fine-tuning \fit scores anticipate post-fine-tuning behavior closely enough to guide model selection and avoid unnecessary tuning?
\end{itemize}

In addressing the above research questions (RQs), we make the following contributions:

\begin{itemize}[noitemsep,leftmargin=*]
    \item Addressing RQ1, we propose \fit, a diagnostic decomposition of cybersecurity QA suitability into vocabulary, knowledge, and contextualization, with a metric for each.
    \item Addressing RQ2, we present an empirical study of how knowledge-focused and instruction-focused fine-tuning trade off these capabilities in small (7B) LLMs, including an abstention-driven inversion of knowledge rankings that we quantify with rank-correlation analysis.
    \item Addressing RQ3, we show that retrieval-grounded contextualization is robust to fine-tuning, and derive practical guidance for model selection in dynamic, low-data domains such as cybersecurity.
\end{itemize}

The rest of the paper is organized as follows. Section \ref{section_objective} formulates the problem. Section \ref{section_background} reviews background and related work. Section \ref{section_methodology} details the \fit tasks. Sections \ref{section_experiment} and \ref{sec:findings} present the experiment, findings, and limitations, followed by concluding remarks.

% ############################################################################################################
\section{Problem Formulation} \label{section_objective}
In this section we define the problem and its foundations. We first describe the implementation scope, then the evaluation tasks; Table~\ref{tab:accents} summarizes the notation used throughout. Fig.~\ref{Fig:task_example} provides a visual reference for the scope and tasks.

\begin{table}[ht]
    \renewcommand{\arraystretch}{1.2}
    \centering
    \caption{Description of Notations.}
    \label{tab:accents}
    \footnotesize
    \begin{tabular}{c|l}
    \hline
    \textbf{Notation} & \textbf{Description}\\
    \hline
    $\prompt$ & {User Input Prompt} \\
    $\{\ner^d \in \ner\}$ & {Domain-specific Vocabulary} \\
    $\{\knowledge^d \in \knowledge \}$ & {Domain-specific Knowledge} \\
    $\{\glb_i^d \mid \glb_i^d \in \glb^d\}$ & {Domain Information for $\prompt$} \\
    $\{\loc_i^d \mid \loc_i^d \in \loc^d\}$ & {Specific Information for $\glb_i^d \cup \prompt$} \\
    $\completion$ & {Completion for $\prompt$ given $(\glb_i^d \cup \loc_i^d) \mid \knowledge^d$ }\\
    $\overline{\ner_i^d},\ \overline{\completion_i}$ & {Ground-truth (expected) outputs} \\
    $\llm$ & {LLM under evaluation} \\
    $\Phi(\cdot \mid \prompt)$ & {Task suitability score over $\prompt$} \\\hline
    \end{tabular}
\end{table}

In a typical knowledge-intensive, critical-domain QA task using an LLM with RAG, the objective is to generate a relevant completion ($\completion$) for a given prompt ($\prompt$) without disclosing sensitive information. Irrespective of the deployment domain, two types of information are primarily required to generate $\completion$. One is domain-specific information ($\glb_i^d$) relevant to $\prompt$; the other is contextual or specific information ($\loc_i^d$) needed to contextualize $\glb_i^d$ for $\prompt$. The LLM then combines its domain vocabulary ($\ner^d$) and knowledge ($\knowledge^d$) to produce $\completion$. To assess an LLM's contextualization ability within this scope, we adopt a process-oriented decomposition into three tasks, each addressing a distinct capability:
\begin{enumerate}
    \item \textbf{Vocabulary.} We assess familiarity with domain vocabulary via a keyword-recognition task that instructs the LLM to identify important keywords ($\ner_i^d$) in $\prompt$.\\
    \item \textbf{Knowledge.} We assess domain knowledge via a multiple-choice QA task probing the LLM's domain understanding ($\knowledge^d$).\\
    \item \textbf{Contextualization.} We assess whether the LLM can comprehend and tailor $\glb_i^d$ in light of $\loc_i^d$ to generate $\completion$ for $\prompt$, without leaking unnecessary information.
\end{enumerate}

This decomposition lets us characterize an LLM's suitability for critical-domain QA from a relevancy and reliability standpoint. We further analyze how each capability changes after fine-tuning, in order to identify systematic patterns of behavioral change. As we show, these patterns are regime-dependent, and understanding them helps practitioners decide whether and how to fine-tune a given model and curate data accordingly.

% ############################################################################################################
\section{Preliminaries}\label{section_background}
The application of pre-trained LLMs in specialized domains has been an active research area \cite{ranade2021generating}. We briefly review the prerequisite background and related developments.

\subsection{LLM, RAG, and Fine-tuning}
LLMs have advanced NLP through transformer architectures \cite{vaswani2017attention} that offer remarkable parallelization \cite{min2023recent}. Pre-trained on massive Internet text with large parameter counts, these models exhibit strong learning capabilities, yet they can produce plausible-but-inaccurate predictions and struggle on problems requiring specialized domain knowledge. Reported reasons \cite{wang2023survey} for the failure of general-purpose LLMs in closed domains include a \textit{deficit in domain knowledge} (lack of exposure to a specialized domain), \textit{outdated information} (a training cutoff that omits post-training developments), and \textit{forgetting} (catastrophic forgetting \cite{kirkpatrick2017overcoming} during additional training, where prior knowledge is lost).

To mitigate knowledge deficiency for domain-specific tasks, an additional knowledge-ingestion step is required. The two most common approaches are Retrieval-Augmented Generation (RAG) and fine-tuning. RAG, introduced by Lewis et al. \cite{lewis2020retrieval}, leverages an external knowledge base (a document corpus, a structured database, or any source of domain information) to overcome the knowledge limitations of pre-trained LLMs. Given an input query, the RAG architecture retrieves the most relevant passages and integrates them into the input, supplying the LLM with additional context.

\begin{figure*}[ht]
  \centering
  \includegraphics[width=0.9\textwidth]{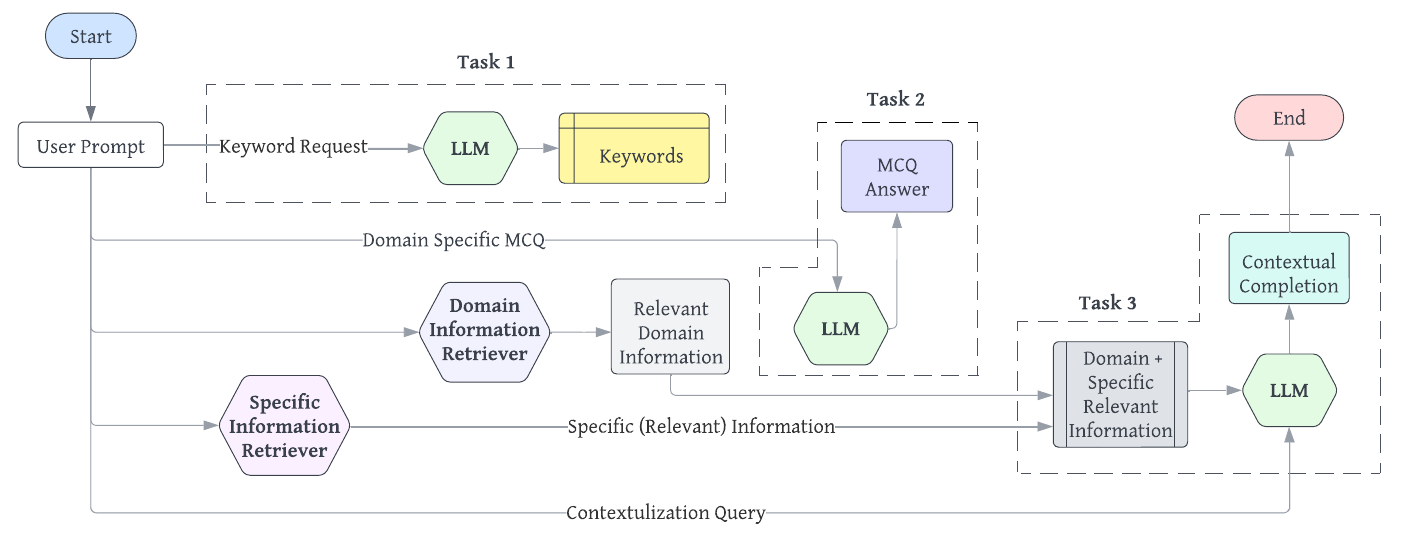}
  \caption{Flowchart of the \fit evaluation process. The three tasks align with the deployment scope: Task 1 assesses vocabulary, Task 2 assesses domain knowledge, and Task 3 assesses contextualization for relevant, reliable generation.}
  \label{Fig:flowchart}
\end{figure*}

As models grow in size, updating all parameters becomes computationally demanding and cost-prohibitive, particularly under limited hardware. This motivates parameter-efficient tuning methods that target strong task performance while minimizing the number of tunable parameters. Representative efforts include \textit{adapter-based} \cite{houlsby2019parameter}, \textit{prompt-based} \cite{lester2021power}, \textit{LoRA} \cite{valipour2022dylora}, \textit{QLoRA} \cite{dettmers2024qlora}, and \textit{hybrid} \cite{mao2021unipelt} approaches.

\subsection{LLM Benchmarking}
Numerous frameworks evaluate general and domain-specific language models across NLP tasks. Some, such as GLUE-X \cite{yang2022glue} and PromptBench \cite{zhu2023promptbench}, assess general capabilities including robustness to \textit{out-of-distribution} and \textit{adversarial inputs}, while KOLA \cite{yu2023kola} evaluates knowledge and reasoning. Domain-specific benchmarks also exist: MultiMedQA targets medical QA, and MATH \cite{singhal2023large} evaluates mathematical reasoning. In cybersecurity, multi-task benchmarks include CyberBench \cite{liucyberbench} and SecureBERT \cite{aghaei2022securebert} for sentiment analysis and NER. These efforts largely report static, pre-trained performance. \fit is complementary: rather than ranking models on a fixed leaderboard, it decomposes suitability into process-aligned capabilities and uses them to study \emph{how} those capabilities shift under fine-tuning, an aspect existing cybersecurity benchmarks do not address.

% ############################################################################################################
\section{\fit Framework}\label{section_methodology}
We now describe the three \fit tasks used to characterize a small LLM's suitability for domain-specific QA with RAG, exemplified in Fig.~\ref{Fig:flowchart}.

\subsection{Task 1: Vocabulary Assessment}
Ensuring a model understands domain vocabulary is a prerequisite for domain-specific QA: it lets the model comprehend the input and respond in compatible terms. We assess this via a \textit{Keyword Recognition (KR)} task, an NLP task that identifies important entities in unstructured text. The LLM is instructed to identify the keywords in the input prompt ($\prompt$). This probes two aspects: the number of correct keyword identifications (a proxy for domain-oriented linguistic understanding, since vocabulary differs sharply across domains, e.g., medical versus cybersecurity), and recognition of instruction-relevant terms (a proxy for understanding the task itself). Formally, let $\prompt = \{\prompt_i\}_{i=1}^n$ be the set of prompts, $d$ the domain, and $\ner^d$ the domain vocabulary. For each $\prompt_i$ there exists a gold keyword set $\overline{\ner_i^d}$. The vocabulary score $\Phi(\ner^d \mid \prompt)$ is

\begin{equation}
   \Phi(\ner^d \mid \prompt) = \frac{1}{n}\sum_{i=1}^n \Delta\!\left(\llm(\prompt_i),\, \overline{\ner_i^d}\right)
\end{equation}

where $\Delta(\cdot,\cdot)$ is the $F_1$ score between the predicted and gold keyword sets, $\llm$ is the model, and $n > 0$.

\begin{tcolorbox}[enhanced,attach boxed title to top center={yshift=-3mm,yshifttext=-1mm},
    colback=white,colframe=gray!75!black,colbacktitle=gray!80!black, title=Task-1: Vocabulary Assessment ($\ner^d$),
      boxed title style={size=small,colframe=gray!90!black}, left=0.5mm, right=0.5mm, boxrule=0.75pt ]
      \small
      \textcolor{black}{\textbf{Instruction:}}\\
        \textcolor{teal}{Print the keywords from the following ...\\}\\
      \textbf{$\prompt_i$}: Can langchain vulnerability affect my system?\\
      \textbf{$\overline{\ner_i^d}$}: \textcolor{green!45!black}{langchain, vulnerability, affect, system}
\end{tcolorbox}

\subsection{Task 2: Knowledge Analysis}
Precisely defining ``knowledge'' is a philosophical question beyond our scope; we instead quantify domain knowledge through a multiple-choice question-answering (MCQ) task. If a model comprehends the question, it can select the correct option, providing a measurable proxy for parametric knowledge. Formally, let $\prompt = \{\prompt_i\}_{i=1}^n$ be a set of MCQ problems, each with $m$ options, and let $\gamma_i$ be the correct answer for $\prompt_i$. Writing $\hat{a}_i = \llm(\prompt_i)$ for the model's selected option and $\mathbb{1}[\cdot]$ for the indicator function, the knowledge score is the accuracy

\begin{equation}
   \Phi(\knowledge^d \mid \prompt) = \frac{1}{n}\sum_{i=1}^n \mathbb{1}\!\left[\hat{a}_i = \gamma_i\right],
\end{equation}

where $\hat{a}_i$ is the model's selected option, $\gamma_i$ the correct option, $m$ the number of choices per item, and $n > 0$.

We note in advance that this score conflates two distinct behaviors when a model declines to answer: a wrong selection and an abstention both reduce accuracy. We exploit this distinction in Section \ref{sec:findings}.

\begin{tcolorbox}[enhanced,attach boxed title to top center={yshift=-3mm,yshifttext=-1mm},
    colback=white!10!white,colframe=gray!75!black,colbacktitle=gray!80!black,
    title=Task-2: Knowledge Analysis ($\knowledge^d$),
    boxed title style={size=small,colframe=gray!50!black}, left=0.5mm, right=0.5mm, boxrule=0.75pt]
    \small
    \textcolor{black}{\textbf{Instruction:}}\\
    \textcolor{teal}{Answer the correct choice for the question ...}

    \textcolor{black}{\textbf{$\prompt_i$}}: \textcolor{black}{A hash function guarantees the integrity of a message. It guarantees that the message has not been
    \\1: Replaced\\2: Overviewed\\3: Changed\\4: Violated \\ }

    \textcolor{black}{\textbf{$\gamma_i$}}: \textcolor{green!45!black}{(3) Changed}
\end{tcolorbox}

\subsection{Task 3: Contextualization Analysis}
Contextualization is the ability to understand and generate text based on the relationships among multi-faceted information. For example, \textit{``What potential impact could CVE-2023-3894 have on the integrity of our TOML configuration data?''} requires jointly reasoning over the CVE and the referenced TOML server. In knowledge-intensive QA, RAG bridges the model's knowledge deficiency; contextualization lets the model produce relevant, reliable answers from retrieved multi-faceted information while following instructions. This ability is critical: in scenarios involving recommendation or privacy, domain information must be tailored to the specifics of the request, and without accurate contextualization a model may provide misinformation or surface inappropriate detail. We therefore adopt contextualized RAG \cite{greshake2023not} as our final task. Comparing the generated response with a Subject-Matter-Expert (SME) reference, we report two RAGAS metrics, \emph{answer correctness} and \emph{semantic similarity}, to assess, respectively, how factually and contextually correct the answer is and how closely it tracks the expert reference. Formally, let $\prompt = \{\prompt_i\}_{i=1}^n$ be domain questions, $\glb_i^d$ the domain information and $\loc_i^d$ the specific information for $\prompt_i$, and $\overline{\completion_i}$ the expert answer. The contextualization score is

\begin{equation}
   \Phi(\completion \mid \prompt) = \frac{1}{n}\sum_{i=1}^n \Omega\!\left(\llm(\prompt_i \mid \glb_i^d \cup \loc_i^d),\, \overline{\completion_i}\right)
\end{equation}

where $\Omega(\cdot,\cdot)$ is the RAGAS scoring function (answer correctness or semantic similarity), $\llm(\prompt_i \mid \glb_i^d \cup \loc_i^d)$ is the model's completion conditioned on the retrieved context, and $n > 0$. We are careful not to interpret semantic similarity as a direct measure of information leakage; it captures fidelity to the expert reference, and we treat it as a reliability signal rather than a security guarantee (see Section \ref{sec:findings}).

\begin{tcolorbox}[enhanced,attach boxed title to top center={yshift=-3mm,yshifttext=-1mm},
    colback=white!10!white,colframe=gray!75!black,colbacktitle=gray!80!black,
    title=Task-3: Contextualization ($\completion$),
    boxed title style={size=small,colframe=gray!50!black}, left=0.5mm, right=0.5mm, boxrule=0.75pt]
    \small
    \textcolor{black}{\textbf{Instructions:}}\\
    \textcolor{teal}{Given the following retrieved knowledge, answer ...}

    \textcolor{black}{\textbf{$\glb_i^d$:}\\}
    \textcolor{teal}{CVE ID: CVE-2023-36189\\
    Description: SQL injection vulnerability in langchain before v0.0.247 allows a remote attacker to obtain ...\\
    CVE ID: CVE-2023-36188\\
    Description: An issue in langchain v.0.0.64 allows a remote attacker to execute arbitrary code via the PALChain ...}

    \textcolor{black}{\textbf{$\loc_i^d$:}\\}
    \textcolor{teal}{`langchain==0.0.270`: Used for generation of document embeddings for text, enhancing information retrieval capabilities ...}

    \textcolor{black}{\textbf{$\prompt_i$}}: \textcolor{black}{Can langchain vulnerability affect my system?}

    \textcolor{black}{\textbf{$\overline{\completion_i}$}}: \textcolor{green!45!black}{Langchain has multiple reported vulnerabilities ... your system is running langchain version 0.0.270, which is not mentioned to be vulnerable in the provided information...}
\end{tcolorbox}

% ############################################################################################################
\section{Experiment \& Evaluation} \label{section_experiment}
We describe the dataset, infrastructure, and evaluation protocol. Cybersecurity, an information-critical domain, serves as our case study.

\begin{table*}[]
\caption{\fit results for pre-trained, knowledge-focused (Finetuned-1), and instruction-focused (Finetuned-2) models. Task~1: keyword-recognition $F_1$; Task~2: MCQ accuracy; Task~3: RAGAS semantic similarity (Sim) and answer correctness (Cor). Cell shading indicates change relative to the pre-trained value (red: decrease, green: increase). Values are from a single fine-tuning run per regime and should be read as indicative (see Section~\ref{sec:findings}).} \label{tab:evaluation}
\footnotesize
\centering
{\renewcommand{\arraystretch}{1.20}%
\begin{tabular}{lrrrr|rrrr|rrrr}
\hline
\multicolumn{1}{l|}{\multirow{4}{*}{\textbf{Model}}} & \multicolumn{4}{c|}{\textbf{Pretrained}}                                                                                & \multicolumn{4}{c|}{\textbf{Finetuned-1}}                                                                               & \multicolumn{4}{c}{\textbf{Finetuned-2}}                                                                               \\ \cline{2-13} 
\multicolumn{1}{l|}{}                       & \multicolumn{1}{c|}{Task 1} & \multicolumn{1}{c|}{Task 2} & \multicolumn{2}{c|}{Task 3}                        & \multicolumn{1}{c|}{Task 1} & \multicolumn{1}{c|}{Task 2} & \multicolumn{2}{c|}{Task 3}                        & \multicolumn{1}{c|}{Task 1} & \multicolumn{1}{c|}{Task 2} & \multicolumn{2}{c}{Task 3}                        \\ \cline{2-13} 
\multicolumn{1}{l|}{}                       & \multicolumn{1}{c|}{F1}     & \multicolumn{1}{c|}{Acc}  & \multicolumn{1}{c|}{Sim}    & \multicolumn{1}{c|}{Cor} & \multicolumn{1}{c|}{F1}     & \multicolumn{1}{c|}{Acc}  & \multicolumn{1}{c|}{Sim}    & \multicolumn{1}{c|}{Cor} & \multicolumn{1}{c|}{F1}     & \multicolumn{1}{c|}{Acc}      & \multicolumn{1}{c|}{Sim}    & \multicolumn{1}{c}{Cor} \\ \cline{1-13} 

\multicolumn{1}{l|}{gpt-3.5-turbo} 
& \multicolumn{1}{r|}{0.85}   & \multicolumn{1}{r|}{0.76}   & \multicolumn{1}{r|}{0.92} & 0.77                   & \multicolumn{1}{r|}{--}   & \multicolumn{1}{r|}{--}   & \multicolumn{1}{r|}{--} & --                   & \multicolumn{1}{r|}{--}   & \multicolumn{1}{r|}{--}   & \multicolumn{1}{r|}{--} & --                   \\ \cline{2-13}

\multicolumn{1}{l|}{llama2-7b} 
& \multicolumn{1}{r|}{0.62}   & \multicolumn{1}{r|}{0.51}   & \multicolumn{1}{r|}{0.91} & 0.78                   & \multicolumn{1}{r|}{\cellcolor{red!50}0.35}   & \multicolumn{1}{r|}{\cellcolor{red!35}0.32}   & \multicolumn{1}{r|}{\cellcolor{red!5}0.87} & \cellcolor{red!5}0.75                   & \multicolumn{1}{r|}{\cellcolor{red!35}0.48}   & \multicolumn{1}{r|}{\cellcolor{red!15}0.31}   & \multicolumn{1}{r|}{\cellcolor{green!15}0.92} & \cellcolor{green!5}0.79                   \\ \cline{2-13}

\multicolumn{1}{l|}{mistral-7b} 
& \multicolumn{1}{r|}{0.47}   & \multicolumn{1}{r|}{0.59}   & \multicolumn{1}{r|}{0.90} & 0.72                   & \multicolumn{1}{r|}{\cellcolor{red!35}0.27}   & \multicolumn{1}{r|}{\cellcolor{red!50}0.39}   & \multicolumn{1}{r|}{\cellcolor{red!5}0.86} & \cellcolor{green!10}0.74                   & \multicolumn{1}{r|}{\cellcolor{red!15}0.43}   & \multicolumn{1}{r|}{\cellcolor{red!25}0.18}   & \multicolumn{1}{r|}{\cellcolor{red!10}0.86} & \cellcolor{green!10}0.76                   \\ \cline{2-13}

\multicolumn{1}{l|}{prometheus-7b} 
& \multicolumn{1}{r|} {0.76}   & \multicolumn{1}{r|}{0.75}   & \multicolumn{1}{r|}{0.92} & 0.73                  & \multicolumn{1}{r|}{\cellcolor{red!15}0.65}   & \multicolumn{1}{r|}{\cellcolor{red!27}0.61}   & \multicolumn{1}{r|}{\cellcolor{red!15}0.85} & 0.73                   & \multicolumn{1}{r|}{\cellcolor{red!50}0.55}   & \multicolumn{1}{r|}{\cellcolor{red!37}0.16}   & \multicolumn{1}{r|}{\cellcolor{red!15}0.86} & \cellcolor{green!20}0.75                   \\ \cline{2-13}

\multicolumn{1}{l|}{westlake-7b}  
                               & \multicolumn{1}{r|}{0.77}   & \multicolumn{1}{r|}{0.71}   & \multicolumn{1}{r|}{0.92} & 0.74                   & \multicolumn{1}{r|}{\cellcolor{red!20}0.59}   & \multicolumn{1}{r|}{\cellcolor{red!15}0.65}   & \multicolumn{1}{r|}{\cellcolor{red!15}0.85} & \cellcolor{red!5}0.73                   & \multicolumn{1}{r|}{\cellcolor{red!45}0.73}   & \multicolumn{1}{r|}{\cellcolor{red!60}0.08}   & \multicolumn{1}{r|}{\cellcolor{green!15}0.93} & \cellcolor{green!40}0.79                   \\ \cline{2-13}

\multicolumn{1}{l|}{westseverus-7b}         & \multicolumn{1}{r|}{0.74}   & \multicolumn{1}{r|}{0.72}   & \multicolumn{1}{r|}{0.89} & 0.69                   & \multicolumn{1}{r|}{\cellcolor{red!5}0.69}   & \multicolumn{1}{r|}{\cellcolor{red!15}0.66}   & \multicolumn{1}{r|}{0.89} & \cellcolor{green!30}0.76                   & \multicolumn{1}{r|}{\cellcolor{red!5}0.73}   & \multicolumn{1}{r|}{\cellcolor{red!50}0.12}   & \multicolumn{1}{r|}{\cellcolor{green!25}0.91} & \cellcolor{green!50}0.78                   \\ \cline{1-13}
\end{tabular}}
\end{table*}

\subsection{Data Description and Preparation}
We construct one dataset per task. \textbf{Task 1} uses 50 cybersecurity questions paired with expert-annotated gold keywords. \textbf{Task 2} uses the computer-security subset of MMLU \cite{mmlu} (100 MCQ items). \textbf{Task 3} draws on two repositories: a domain information repository ($\glb^d$) built from NIST \cite{nist}, and a QA-specific information repository ($\loc^d$) curated from an organization-specific infrastructure wiki; since such infrastructure detail is sensitive, we substitute synthetic data of the same form. We then author 50 questions that require both repositories to answer, each with an SME ground-truth completion. For fine-tuning we prepare two datasets: a knowledge-focused set (Finetuned-1), built by generating QA pairs from the Cisco Talos corpus \cite{talos}, and an instruction-focused set (Finetuned-2), drawn from a training split of the Task-3 evaluation data with explicit instructions to abstain when uncertain. All curated datasets and fine-tuned checkpoints will be released. We report dataset sizes explicitly since, given the modest $n$, individual numeric differences should be read as indicative rather than significant (Section~\ref{sec:findings}).

\subsection{Experiment Infrastructure}
We evaluate five open-weight 7-billion-parameter, 4-bit-quantized QA models: \textit{Llama-2-7b\footnote{huggingface.co/meta-llama/Llama-2-7b-chat-hf}, Mistral-7b\footnote{huggingface.co/mistralai/Mistral-7B-Instruct-v0.2}, Prometheus-7b\footnote{huggingface.co/AiMavenAi/AiMaven-Prometheus}, WestLake-7b\footnote{huggingface.co/senseable/WestLake-7B-v2}, and WestSeverus-7b\footnote{huggingface.co/FelixChao/WestSeverus-7B-DPO-v2}}. We also include \textit{GPT-3.5-Turbo\footnote{platform.openai.com/docs/models/gpt-3-5-turbo}} as a strong reference point (pre-trained only; we do not fine-tune the API model). These models were chosen deliberately: their pre-training predates much of the evolving threat intelligence (recent CVEs, advisories, infrastructure detail) used here for fine-tuning and evaluation, which reduces data-contamination risk and lets us attribute gains to retrieval and fine-tuning rather than prior exposure. For retrieval we use \textit{ChromaDB\footnote{trychroma.com}} as the vector store. Experiments ran on an \textit{Intel i9-12900 with an NVIDIA GeForce RTX\texttrademark{} 3090\,Ti} and 128\,GB RAM. All models were decoded greedily (temperature $0$) for deterministic, comparable outputs; the 4-bit quantization is held fixed across models so comparisons are like-for-like, though quantization may shift absolute scores.

\subsection{Evaluation Protocol}
We combine quantitative and qualitative evaluation. The quantitative evaluation uses the RAGAS \cite{es2023ragas} framework for Task 3; the qualitative evaluation uses two cybersecurity SMEs to judge generated responses. Results appear in Tables \ref{tab:evaluation}--\ref{tab:felis}.

\subsubsection{Quantitative Evaluation}
For Task 1 (KR) we report $F_1$ between predicted and gold keywords. Task 2 (MCQ) is scored by accuracy. For Task 3 we report RAGAS answer correctness and semantic similarity. We use RAGAS rather than BLEU \cite{papineni2002bleu} or ROUGE \cite{rouge2004package} because the latter are tailored to machine translation and summarization and correlate poorly with answer correctness in QA.

\subsubsection{Qualitative Evaluation}
% \begin{table}[ht]
% \renewcommand{\arraystretch}{1.2}
% \centering
% \caption{Inter-rater agreement (Fleiss' $\kappa$) between the two cybersecurity SMEs.}
% \label{tab:felis}
% \footnotesize
% \begin{tabular}{l|c}
% \toprule
% \textbf{Model} & \textbf{Fleiss' $\kappa$}\\
% \midrule
% gpt-3.5-turbo  & 0.861\\
% llama2-7b      & 0.845\\
% mistral-7b     & 0.782\\
% prometheus-7b  & 0.864\\
% westlake-7b    & 0.944\\
% westseverus-7b & 0.868\\
% \bottomrule
% \end{tabular}
% \end{table}

\begin{table}[]
\footnotesize
\centering
{\renewcommand{\arraystretch}{1.20}%
\caption{Inter-rater agreement (Fleiss Kappa) between the two cybersecurity SMEs across models.} \label{tab:felis}
\centering
\begin{tabular}{l|r|r}

\hline
\multicolumn{1}{c|}{\textbf{Model}} & \multicolumn{1}{c|}{\textbf{Kappa ($K$)}} & \multicolumn{1}{c}{\textbf{Standard Error}} \\ \hline
gpt-3.5-turbo               & \cellcolor[HTML]{D6F8D6}0.861  & 0.080                       \\ \hline
llama2-7b                   & \cellcolor[HTML]{D1F1D1}0.845  & 0.084                      \\ \hline
mistral-7b                  & \cellcolor[HTML]{EFFEEF}0.782  & 0.082                      \\ \hline
prometheus-7b               & \cellcolor[HTML]{94F894}0.864  & 0.077                      \\ \hline
westlake-7b                 & \cellcolor[HTML]{34FF34}0.944  & 0.081                      \\ \hline
westseverus-7b              & \cellcolor[HTML]{91F791}0.868  & 0.078                      \\ \hline
\end{tabular}}
\end{table}

Two SMEs assessed \fit's contextual responses on a 5-point Likert scale \cite{allen2007likert}, from 1 (``factually incorrect and contextually irrelevant'') to 5 (``factually accurate and contextually relevant''). Inter-rater agreement, measured by Fleiss' $\kappa$ \cite{mchugh2012interrater} (Table \ref{tab:felis}), was strong for most models (gpt-3.5-turbo 0.861, llama2-7b 0.845, prometheus-7b 0.864, westlake-7b 0.944, westseverus-7b 0.868) and moderate for mistral-7b (0.782). This agreement establishes the reliability of the SME judgments used to ground the Task-3 references.

\subsection{Fine-tuning}
We fine-tune with QLoRA \cite{dettmers2024qlora}, a PEFT \cite{ding2023parameter} method, under both the knowledge-focused and instruction-focused regimes. Hyper-parameters were held constant across regimes (rank 64, batch size 4, 5 epochs). Each regime was run once per model; we therefore frame the resulting numbers as indicative and analyze \emph{patterns} of change rather than individual cell differences.

% ############################################################################################################
\section{Findings and Limitations} \label{sec:findings}

\subsection{Fine-tuning does not uniformly help small models}
Across Table \ref{tab:evaluation}, both fine-tuning regimes degrade vocabulary (Task 1) and parametric knowledge (Task 2) for every model relative to its pre-trained baseline. In a domain like cybersecurity, where prompts routinely contain newly disclosed terms and CVEs, knowledge-focused tuning (Finetuned-1) cannot keep pace: it lowers keyword $F_1$ for all models (e.g., Llama-2 $0.62\!\rightarrow\!0.35$) and reduces MCQ accuracy in parallel. The practical implication is that, absent abundant and current labeled data, a well-chosen pre-trained model paired with RAG is often the safer choice.

\subsection{The two regimes trade off differently but predictably in direction}
\begin{table}[ht]
\renewcommand{\arraystretch}{1.2}
\centering
\caption{Spearman rank correlation ($\rho$) between pre-trained and post-fine-tuning scores across the five open-weight models. High positive $\rho$ indicates preserved rankings; negative $\rho$ indicates inversion.}
\label{tab:rankcorr}
\footnotesize
\begin{tabular}{l|cc}
\toprule
\textbf{Task} & \textbf{Pre $\rightarrow$ Finetuned-1} & \textbf{Pre $\rightarrow$ Finetuned-2}\\
\midrule
Task 1 (vocabulary) & $0.60$ & $0.82$\\
Task 2 (knowledge)  & $0.70$ & $-0.60$\\
Task 3 (correctness)& \multicolumn{2}{c}{not interpretable (range $0.69$--$0.79$)}\\
\bottomrule
\end{tabular}
\end{table}

To quantify how rankings move, we compute Spearman rank correlations between pre-trained and post-tuned scores (Table \ref{tab:rankcorr}). Vocabulary rankings are well preserved under both regimes ($\rho=0.60$ and $0.82$), and knowledge rankings are preserved under knowledge-focused tuning ($\rho=0.70$). Strikingly, knowledge rankings \emph{invert} under instruction-focused tuning ($\rho=-0.60$): the strongest pre-trained knowledge models (WestLake, WestSeverus) become the weakest on Task 2 (accuracy $0.08$ and $0.12$). This inversion is the key reason a naive ``best-before-equals-best-after'' heuristic fails, and why \fit must be read as a regime-aware diagnostic rather than a monotonic predictor.

\subsection{The instruction-tuning collapse is abstention, not knowledge loss}
The Task-2 collapse under Finetuned-2 is an artifact of the abstention instruction, not erasure of knowledge. Two pieces of evidence support this. First, the instruction-focused data explicitly directed the model to withhold an answer when uncertain, making conservatism the trained behavior. Second, and more tellingly, the same instruction-tuned models retain or slightly \emph{improve} contextual answer correctness (Task 3, e.g., WestLake $0.74\!\rightarrow\!0.79$) even as their standalone MCQ accuracy falls to near zero. A model that has truly lost knowledge could not answer correctly when that knowledge is retrieved; these models can. Instruction tuning thus trades parametric recall for caution, a property that is harmful for closed-book MCQ but potentially desirable for retrieval-grounded deployment, where unsupported answers are a liability. This aligns with the abstention behavior described in \cite{xin2021art}.

\subsection{Retrieval-grounded contextualization is robust}
Task-3 correctness is remarkably stable across all models and conditions (range $0.69$--$0.79$), and semantic similarity is uniformly high ($\sim\!0.85$--$0.93$). When relevant context is retrieved and supplied, model choice and fine-tuning have little effect on contextual correctness. We therefore caution against interpreting similarity as an information-leakage metric: it is saturated and non-discriminative here, and reflects fidelity to the expert reference rather than the presence or absence of sensitive disclosure. The practical takeaway is that, for contextual correctness in this setting, investment in retrieval quality yields more than investment in fine-tuning.

\subsection{Selection guidance}
Taken together, these patterns make \fit useful as a pre-adaptation screen: weak pre-trained models (e.g., Mistral on vocabulary) remain weak after tuning, and the strongest knowledge models survive knowledge-focused tuning. Where instruction-focused tuning is planned, practitioners should expect, and can pre-empt, an abstention-driven collapse in closed-book knowledge, and should evaluate such models in their retrieval-grounded configuration rather than on standalone knowledge probes.

\subsection{Threats to Validity}
\textbf{Construct.} Our metrics are proxies. Task~2 accuracy conflates a wrong answer with a deliberate abstention; we cross-check against retrieval-grounded correctness (Task~3), but the proxy is imperfect. RAGAS similarity captures fidelity to the expert reference, not information leakage, so we read it only as a reliability signal. \textbf{Internal.} Each regime was run once with fixed hyper-parameters and greedy decoding, so regime effects are not separated from run-to-run variance; Finetuned-1 uses Cisco Talos while Task~2 uses MMLU, so part of the Task~2 drop may be distribution shift rather than a general fine-tuning effect; fixed 4-bit quantization keeps comparisons like-for-like but may shift absolute scores; and using models whose pre-training predates the evaluation content reduces, but does not eliminate, contamination risk. \textbf{External.} We cover five open-weight models at one 7B scale plus an API reference in a single domain, so patterns may not transfer across scales, architectures, or domains \cite{liucyberbench}; Task-3 specific data are synthetic for confidentiality and the threat-intelligence snapshot is fixed in time. \textbf{Conclusion.} Datasets are modest ($n=50/100/50$) and we report no significance tests, so numeric differences are indicative; in particular, the Spearman values in Table~\ref{tab:rankcorr} are computed over five models and are descriptive of direction rather than statistically significant. A larger model pool and repeated runs would be needed to test these trends formally.

% ############################################################################################################
\section{Conclusion}
Cybersecurity is tied to critical infrastructure, making the reliable deployment of LLMs in this domain consequential. We presented \fit, a task-oriented diagnostic that decomposes cybersecurity QA suitability into vocabulary, knowledge, and contextualization, and used it to study how five small LLMs behave under knowledge-focused and instruction-focused fine-tuning. Fine-tuning did not uniformly help: it degraded parametric capabilities in every model, knowledge-focused tuning preserved relative rankings, and instruction-focused tuning collapsed measured knowledge through abstention while leaving retrieval-grounded contextualization intact, a regime-dependent trade-off we quantified with rank-correlation analysis. These results indicate that, in dynamic, low-data domains, a well-chosen pre-trained model paired with strong retrieval often strikes a better balance than fine-tuning, and that pre-adaptation diagnosis can guide model selection and reduce unnecessary tuning cost.

% ############################################################################################################
\section*{Ethics Statement}
Our study uses datasets that contain no sensitive information. To obtain cyber-threat intelligence, specifically Common Vulnerabilities and Exposures (CVEs), we used web crawlers issuing API calls within the limits set by authorized sources. We anonymized our human evaluators and ensured no personally identifiable information was disclosed. We confirm that our research aligns, to the best of our knowledge, with the IEEE Code of Ethics.

\section*{Acknowledgment}
This work was supported by the National Science Foundation under Grant No.~2611682. Any opinions, findings, and conclusions or recommendations expressed in this material are those of the author(s) and do not necessarily reflect the views of their institution or the National Science Foundation.

% ############################################################################################################
% Appendix figures (span both columns)
% \appendices
% \section{Per-Model Task Breakdowns}
% For completeness, Figs.~\ref{Fig:pretrained}, \ref{Fig:finetune1}, and \ref{Fig:finetune2} provide the per-model breakdowns of Task~1 and Task~3 used in Section~\ref{sec:findings}, for the pre-trained, knowledge-focused (Finetuned-1), and instruction-focused (Finetuned-2) settings, respectively.

\begin{figure*}[p]
\centering
\includegraphics[height=0.265\textheight,keepaspectratio, trim=1in 5in 1in 0in, clip]{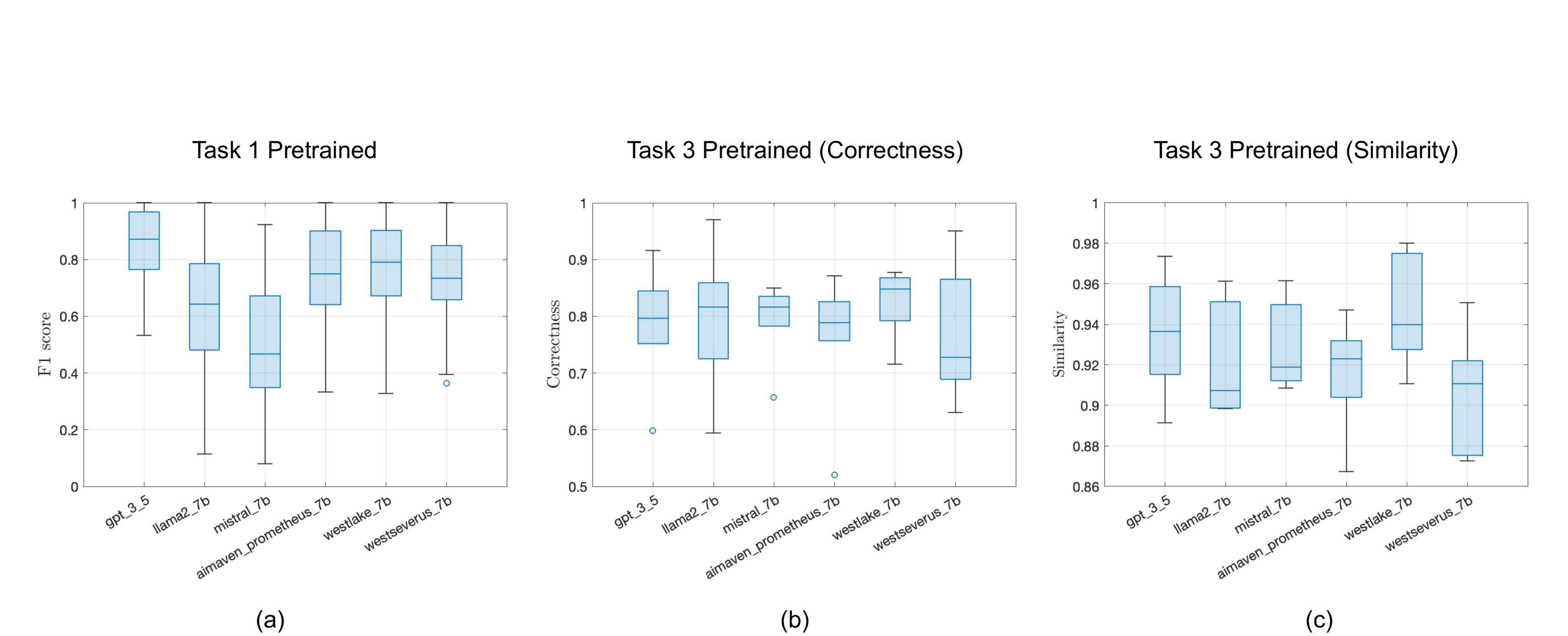}
\caption{Performance of pre-trained models on Tasks 1 and 3. (a) $F_1$ on vocabulary assessment; (b) answer correctness on contextual completion; (c) semantic similarity of completions.}
\label{Fig:pretrained}

\vspace{0.6em}
\includegraphics[height=0.265\textheight,keepaspectratio, trim=1.5in 5in 1in 1.5in, clip]{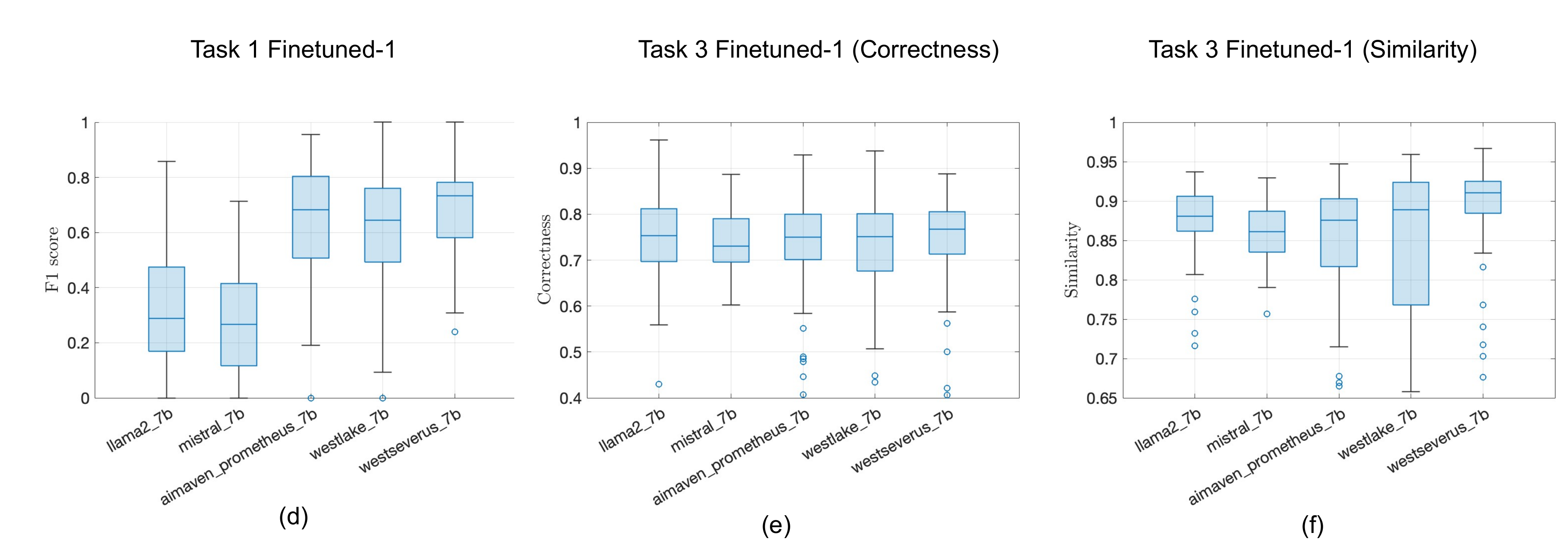}
\caption{Performance of Finetuned-1 (knowledge-focused) models on Tasks 1 and 3. (a) $F_1$ on vocabulary assessment; (b) answer correctness on contextual completion; (c) semantic similarity of completions.}
\label{Fig:finetune1}

\vspace{0.6em}
\includegraphics[height=0.265\textheight,keepaspectratio, trim=2in 5in 1in 1.5in, clip]{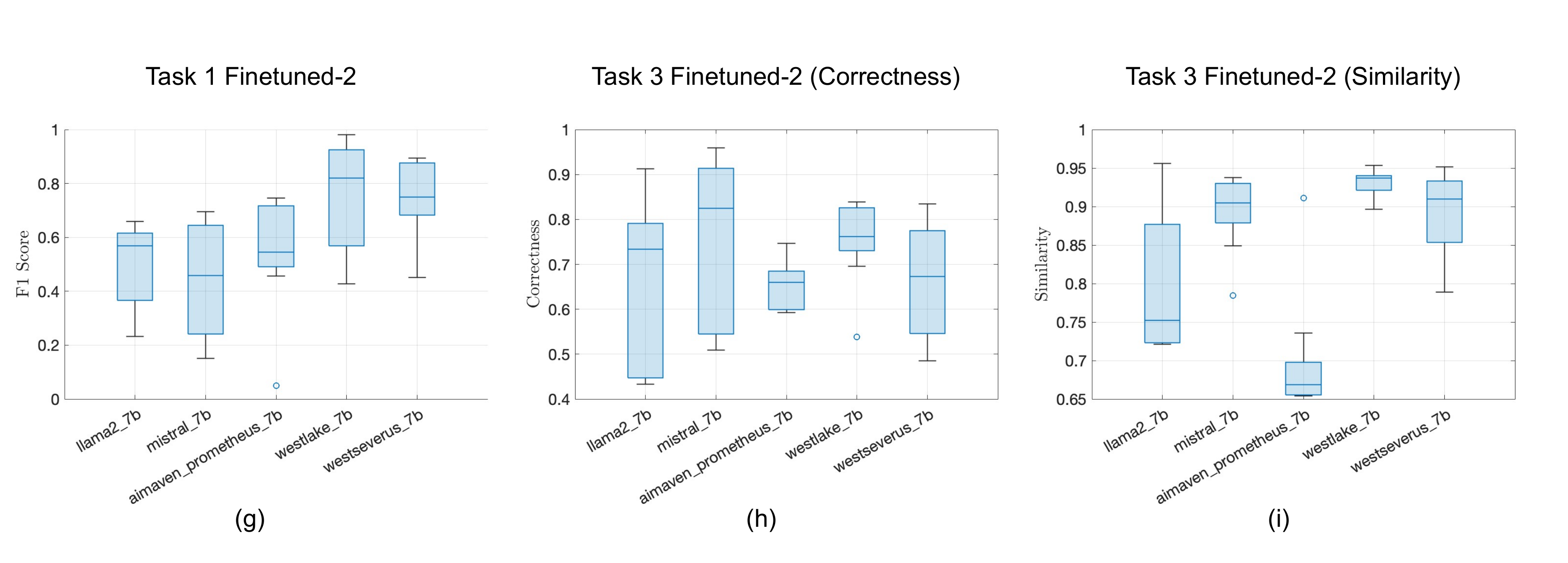}
\caption{Performance of Finetuned-2 (instruction-focused) models on Tasks 1 and 3. (a) $F_1$ on vocabulary assessment; (b) answer correctness on contextual completion; (c) semantic similarity of completions.}
\label{Fig:finetune2}
\end{figure*}

\balance
\bibliographystyle{ieeetr}
\bibliography{references}

\end{document}